# From Maxout to Channel-Out: Encoding Information on Sparse Pathways

Qi Wang
Department of ECE
and Institute for Advanced Computer Studies
University of Maryland
College Park, MD
qwang37@umiacs.umd.edu

Joseph JaJa
Department of ECE
and Institute for Advanced Computer Studies
University of Maryland
College Park, MD
joseph@umiacs.umd.edu

## Abstract

*Motivated by an important insight from neural science, we propose a new framework for understanding the success of the recently proposed "maxout" networks. The framework is based on encoding information on sparse pathways and recognizing the correct pathway at inference time. Elaborating further on this insight, we propose a novel deep network architecture, called "channel-out" network, which takes a much better advantage of sparse pathway encoding. In channel-out networks, pathways are not only formed a posteriori, but they are also actively selected according to the inference outputs from the lower layers. From a mathematical perspective, channel-out networks can represent a wider class of piece-wise continuous functions, thereby endowing the network with more expressive power than that of maxout networks. We test our channel-out networks on several well-known image classification benchmarks, setting new state-of-the-art performance on CIFAR-100 and STL-10, which represent some of the "harder" image classification benchmarks.*

## 1. Introduction

Most of the recent work on deep learning has focused on ways to regularize network behavior to avoid over-fitting. Dropout [8] has been widely accepted as an effective way for deep network regularization. Dropout was initially proposed to avoid co-adaptation of feature detectors, but it turns out it can also be regarded as an efficient ensemble model. The maxout network [7] is a newly proposed micro architecture of deep networks, which works well with the dropout technique. It sets the state-of-the-art performance on many popular image classification datasets. In retrospect, both methods follow the same approach: they restrict updates triggered by a training sample to affect only a sparse sub-graph of the network. We note that neural science researchers have come up with what can be perceived as a similar principle from years of study of the human brain: It is not the shape of the signal, but the pathway along which the signal flows, that determines the functionality [10]. We use this principle as an important clue to explore the encoding capabilities of deep networks.

In this paper we provide a new insight into a possible reason for the success of maxout, namely that it partially takes advantage of what we call "sparse pathway coding", a much more robust way of encoding categorical information than encoding by magnitudes. In sparse pathway encoding, the pathway selection itself carries significant amount of the categorical information. With a carefully designed scheme, the network can extract pattern-specific pathways during training time and recognize the correct pathway at inference time. Guided by this principle, we propose a new class of network architectures called "channel-out networks". Unlike the maxout network, this type of networks does not only form sparse input pathways a posteriori, but it also actively selects outgoing pathways. We run experiments on channel-out networks using several image classification benchmarks, all of which show similar or better performance results compared with maxout. In fact, the channel-out network sets new state-of-the-art performance on some image classification datasets that are on the "harder" end of the spectrum (i.e., those with lower current state-of-the-art performance), namely CIFAR-100 and STL-10, demonstrating its potential to encode large amounts of information with higher level of complexity. Channel-out is just one example of using the sparse pathway encoding principle, which we believe is a promising research direction for designing other types of deep network architectures.

## 2. Dropout vs Maxout: Ensemble of Sub-Models vs Recognizable Sub-Models

In this section we briefly review the concepts of dropout [8] and maxout [7], and introduce the concept of sparse pathway encoding.

Dropout regularizes the network during the training phase by randomly crossing out some of the nodes upon the processing of each training sample. Therefore it restricts the updates to happen only along a relatively sparse subnetwork. As has been pointed out in a number of papers [8, 18, 20], dropout networks can be regarded as performing efficient model averaging of a large ensemble of models randomly sampled from the original network, so that each model learns a different representation of the data.

The maxout network is a recently proposed microarchitecture that is significantly different from traditional networks: the activation function does not take a normal single-input-single-output form, but instead the maximum of the activations from several candidate nodes. In [7], the advantage of maxout over normal differentiable activation functions (such as tanh) was attributed to its better approximation to the exact model averaging, and the advantage of maxout over the rectified linear activation function was attributed to easier optimization on maxout networks. Here we propose another insight of the power of the maxout network. The maxout node activates only one of the candidate input pathways depending on the input, and when the gradient (the novel information discovered from the current training sample) is back-propagated, only that selected pathway is updated with the information, which it encodes (Figure 1). Therefore the information conveyed by each training sample is compactly encoded onto a local portion of the network, instead of being disseminated across the entire network. We call this behavior as "sparse pathway coding". Moreover, due to the local invariance of the max-linear function, the network can "remember" the correct pathway selection for the same or similar test sample. At inference time, if a test sample is similar to a certain pattern encoded, then it is more likely to activate a pathway that is similar to the one activated by this pattern. Correct inference can therefore be made with awareness of that pattern-specific pathway.

To sum up, The maxout network encodes information sparsely on distinct pathways, and has the capability of retrieving the correct pathway at inference time. This observation is well in line with the famous principle in neural science, "The information conveyed by an action potential is determined not by the form of the signal but by the pathway the signal travels in the brain" [10].

## 3. Channel-Out Networks

### 3.1. Pushing the sparse pathway concept further

Although maxout networks implement the concept of sparse pathway coding, the pathways are formed "a posteriori", meaning that a pathway selection is only implicitly implemented with already-formed input paths, but no active link selection is ever made. Higher layer pathway selections are not aware of the selections made in lower layers. A natural extension is to endow the network with the capability of active output pathway selection, which may result in a more efficient pathway coding. Along this line, we propose a class of new deep network architectures called "channel-out networks".

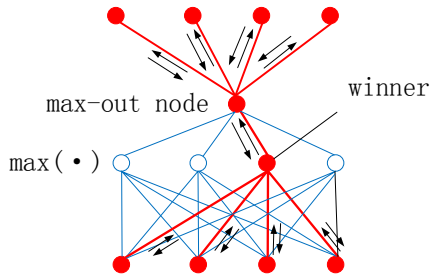

Figure 1: Sparse pathway coding by maxout networks. Effective information flows only along the red links.

A channel-out network is characterized by channel-out groups (Figure 2). At the end of the linear portion in a typical layer (can be fully connected, convolutional, or locally connected), output nodes are arranged into groups, and for each group, a special channel selection function is performed to decide which channel opens for further information flow. Only the activation of the selected channel is passed through, all other channels are blocked off (this is realized by masking the corresponding channel output by 0). When the gradient is back-propagated through this channel-out layer, it only passes through the open channels selected during forward propagation. Formally, we first define a vector-valued channel selection function $f(a_1, a_2, ..., a_k)$ which takes as input a vector of length $k$ and outputs an index set of length $l$ ($l < k$). Elements of the index set are selected from the domain $\{1, 2, ..., k\}$. Then with an input vector (typically the previous layer output) $\mathbf{a} = (a_1, a_2, ..., a_k) \in \mathcal{R}^k$, a channel-out group implements the following activation functions:

$$h_i = \mathbf{I}_{\{i \in f(a_1, a_2, ..., a_k)\}} a_i \quad (1)$$

where $\mathbf{I}(\cdot)$ is the indicator function, $i$ indexes the candidates in the channel-out group, $a_i$ is the $i^{th}$ candidate input, and $h_i$ is the output (Figure 2). There are many possible choices of the channel selection function $f(\cdot)$. To ensure good performance, we require that the channel selection function possesses the following properties:

- The function must be piece-wise constant, and the piece-wise constant regions should not be too small. Intuitively, the function has to be "regular enough" to ensure robustness against the noise in the data.

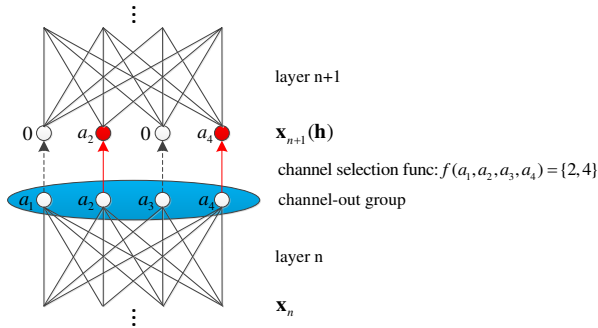

Figure 2: Operation performed by a channel-out group

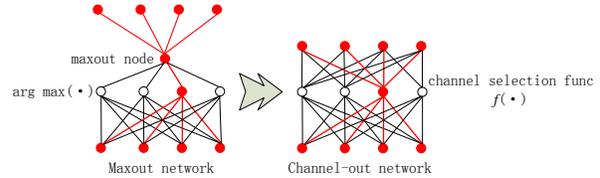

Figure 3: Difference between maxout and channel-out: A maxout node is attached to a set of FIXED output links, resulting in same output pathway for different input pathways; A channel-out group is connected to a set of different output links, resulting in distinct output pathways.

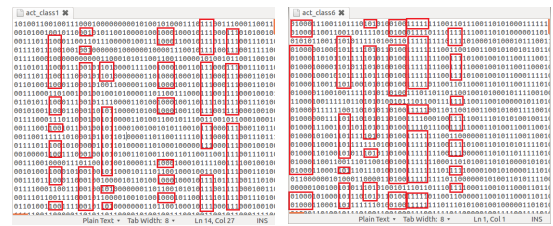

(a) Automobile          (b) Frog

Figure 4: Pathway patterns for two classes in CIFAR-10

- The pre-image size of each possible index output must be of almost the same size. In other words, each channel in the channel-out group should be equally likely to be selected as we process the training examples (so that the information capacity of the network is fully utilized).

- The computation cost for evaluating the function must be as low as possible.

Some examples of good channel selection functions are: $\arg\max(\cdot)$, $\arg\min(\cdot)$, $\arg\mathrm{median}(\cdot)$, indices of the $l$ largest candidates, and the absolute-max $\arg\max(|\cdot|)$. The test results reported in this paper all used the $\arg\max(\cdot)$ function.

The characteristics of a channel-out network is better illustrated through a comparison with the maxout network (Figure 3). In a maxout network, a maxout node only takes the maximum value of the candidates, without the outgoing links being aware of the index. In contrast, in the channel-out network, the channel-out group knows about the index of the open channel, and output link selection is made possible by this extra piece of information. In summary, the main characteristic of a channel-out network is that a channel-out group uses channel selection information to determine future inference pathways.

In a deep convolutional network for 2D image classification, channel-out groups can be implemented by performing the channel selection function across groups of feature maps after convolution/multiplication. Similar to maxout networks, channel-out can be regarded as a special kind of cross-feature pooling. Moreover, in addition to the channel output value, $h_{f(\mathbf{a})}$, the index vector itself $f(\mathbf{a})$ should also be recorded into an index matrix, so that the open pathway can be retrieved during back-propagation. We can further take advantage of the index vector to implement sparse convolution and sparse matrix multiplication, avoiding wasteful computations of multiplications by zeros. This can potentially make the computation very fast. Training a channel-out network that has a similar number of parameters as a maxout network, assuming a scalar channel selection function, can in theory be done $k$ times faster than the maxout network, where $k$ is the size of channel-out/maxout groups. We are currently working on this fast implementation.

To confirm that pathway selection is indeed indicative of the patterns in the data, we record the pathway selections of a well-trained maxout model and a channel-out model (with $\max(\cdot)$ channel selection function) using the CIFAR-10 dataset. For ease of visualization and analysis, we set the size of maxout/channel-out groups to 2, so that we can use 0/1 to represent the pathway selection at each maxout/channel-out group. Figure 4 shows the channel-out pathway patterns of two classes, where each row records the pathway selections of all channel-out groups (only part of them are shown in the window) when performing inference on a specific test sample. We can clearly observe many class-specific string patterns, indicating that pathway selection patterns indeed subsume important clue of categorization. Some of the salient patterns have been marked with red rectangles.

To better visualize the space of pathway patterns, we perform PCA analysis on the pathway pattern vectors and project them into the three dimensional space. Figures 5 and 6 show the results for channel-out and maxout, respectively. We can see that clusters have been well formed.

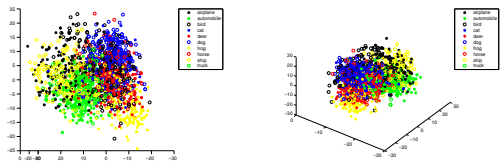

Figure 5: 3D visualization of the pathway pattern: channel-out

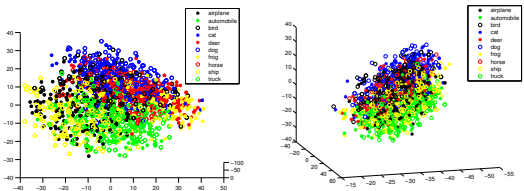

Figure 6: 3D visualization of the pathway pattern: maxout

Although these clusters are not perfect, they still demonstrate that considerable amount of categorical information has been encoded by pathway patterns. We can also see that the channel-out model results in better clusters than those generated by maxout. For example, the frog class (yellow stars) is better separated from the other classes in the channel-out case. Another interesting observation is that, while the channel-out and maxout models have been trained independently with completely random parameter initialization, the relative 3D positions of the classes in the visualization show very similar spatial structures. This implies that the pathway pattern provides a very robust representation of the underlying patterns in the data.

In the next few sub-sections we argue from different perspectives why the channel-out network manifests the concept of sparse pathway selection better than maxout, and why it can potentially exhibit better performances on classification tasks.

### 3.2. Expressive power of the channel-out network

Universal representation/approximation property is a good indication of the expressive power of a model. The fact that a two-layer circuit of logic gates can represent any boolean function [14] implies the nonlinear expressive power of early shallow neural networks. Existence of functions that are computable with polynomial-size networks with $k$ layers while requiring exponential-size networks with $k-1$ layers implies the benefit of using deep architectures [1]. More recently, the fact that maxout networks are universal approximator of any continuous function is an indication of its good performance [7]. Channel-out networks also have strong universal approximation properties. Take the $\max(\cdot)$ channel selection function as an example, we can prove the following theorem:

***Theorem:*** *Any piece-wise continuous function defined on a compact domain in Euclidean space can be approximated arbitrarily well by a $\max(\cdot)$ two-layer channel-out network with one hidden channel-out group with finite number of candidate nodes. That is, denote the target function to be approximated as $T(\mathbf{x})$, then $\forall \epsilon > 0$, we can find a two-layer channel-out network with one hidden channel-out group that implements a function $F(\cdot)$ such that $\int |T(\mathbf{x}) - F(\mathbf{x})|^2 d\mathbf{x} < \epsilon$. Here by a "piece-wise continuous function" we refer to a function that is constituted of finite number of continuous segments.*

The proof is straightforward and is not of much technical significance. It is listed in the Appendix section.

### 3.3. Channel-out network does more efficient pathway switching

Local learning behaviors on maxout and channel-out networks can be categorized into two classes: activation tuning (mild change) and pathway switching (critical change). Activation tuning happens when the desired parameter update is not significant enough to change the pathway selection, otherwise pathway switching occurs. Channel-out and maxout networks show very different behaviors upon pathway switching. We discuss it under a simple scenario in which we repeatedly present the same training sample to the network. For a maxout node, when pathway switching happens, the gradient propagated to the maxout node must be to decrease the activation. However, the effective activation decrease will be thresholded by the activation level of the second largest candidate activation, which slows down, but does not steer away the decreasing trend (we are assuming that the network structure above this maxout node is relatively invariant across different presentations of the training sample). Under this scenario, updates will typically alternate between candidate pathways, reducing the overall efficiency of updates. In contrast, for a channel-out network, whenever a pathway switches in a channel-out group, the effective structure above the channel-out group is drastically changed due to the distinct output links selected by the channel-out group, resulting in significant change in the desired direction of the activation update. In other words, when current pathway is too far from fitting the data well, the channel-out group tends to switch and start training a new pathway from scratch. Clearly this is a more desired behavior for pathway switching.

## 4. Sparse Pathway Selection as a Regularization Method

In this section we consider different principles underlying sparse pathway methods (channel-out/maxout) and dropout in terms of regularizing network training. We argue that the strengths of sparse pathway selection and dropout complement each other. This explains the main reason that sparse pathway methods, when combined with dropout, outperforms traditional neural network models. We believe that sparse pathway encoding is a general direction that is well worth pursuing in future research for designing deep network architectures. Maxout and channel-out networks are just two examples along this direction.

### 4.1. Dropout regularization: squeezing all information into each sub-network

Recall that dropout, with each presentation of a training sample, samples a sub-network and encodes the information revealed by the training sample onto this sub-network. Since the sampling of data and sub-networks are independent processes, in a statistical sense the information provided by each training sample will eventually be "squeezed" into all these sub-networks (third row of Figure 7).

The advantage of such scheme, as has been pointed out in various papers [8, 18, 20], is that the same piece of information is encoded into many different representations, adding to the robustness at inference time. The side-effect, which has not been highlighted before, is that encoding conflicting pieces of information densely into sub-networks with small capacities causes interference problem. Data samples of different patterns (classes) attempt to build different, maybe highly conflicting network representations. When the sub-network is not large enough to hold all the information, conflicting parts tend to cancel each other, resulting in significant information loss. An extreme case is the practice of training several networks independently (on the same dataset) and combine them by naive averaging of the link weights, which clearly does not make much sense in general.

Note that although dropout is disseminating the training set information over the entire network, it is significantly different from a normal deep network without any regularization. For the latter, retrieval of information relies on overall cooperation of the entire network, where each sub-network only encodes PART of the information (second row of Figure 7). In contrast the dropout network tries to encode the entire information in EACH sub-network. Sub-networks interact with each other more by averaging (or voting), rather than by cooperation.

### 4.2. Sparse pathway regularization: encoding different patterns into different sub-networks

In contrast to dropout, sparse pathway regularization methods tend to encode information in a more specialized way. Each pattern of the training data is likely to be encoded onto one or a few specialized sub-networks (for the simplest case, readers can consider each class as of one pattern). This is illustrated in the fourth row of Figure 7.

Clearly, sparse pathway methods mitigate the interference problem caused by dropout. The problem with pure sparse path way regularization, however, is that network capacity could be under-utilized. Since sub-networks are highly pattern-specific, and there are only a few fundamental patterns in the data, only a small subset of the entire network is selected and trained, leaving the rest of the network capacity in an "idle" state. This is a waste of the network capacity. Now it is not hard to understand why combining dropout and sparse pathway encoding could result in better performance: the combination takes advantage of the strengths of both methods to do more efficient information coding, as illustrated in the last row of Figure 7.

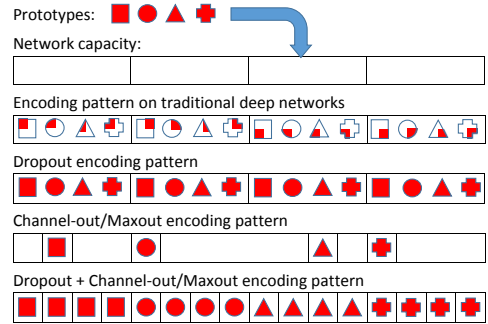

Figure 7: Information encoding patterns. Each bin of the network capacity represents a certain size sub-network. Dropout tends to encode all patterns to each capacity bin, resulting in efficient use of network capacity but high level of interference; Sparse pathway methods tend to encode each pattern to a specific sparse sub-network, resulting in least interference but waste of network capacity; The best approach is the combination of the two schemes.

Notice that encoding data pattern(s) into sparse pathways is also different from encoding partial patterns into sub-networks, as is done in a normal network without regularization. In a non-regularized network, the information encoded on a sub-network is not likely to form complete patterns, while in sparse pathway methods a certain sub-network always tries to encode complete patterns. Here we note also the significant difference between sparse pathway methods and standard sparse feature learning techniques [3, 13, 16], which is that a sparse pathway model must also possess the ability of RECOGNIZING the correct pathway during inference time.

An empirical observation in support of our arguments about dropout/sparse pathway encoding pattern is seen when we take extremes of either of the regularization directions in our experiments. We observe that when dropout rate is set too high (number of alive nodes too few for each training sample), the dominating problem is under-fitting, where we see both training and test set precision cease to improve at very early epochs. When no dropout is applied, the dominating problem is over-fitting, where we see training set precision growing very fast while test set precision is not catching up with the improvement of training set fitness. This is in line with our major assumption about deficiencies of each method: dropout causes interference, while pure sparse pathway encoding forgoes using redundant network capacity to enhance robustness, thus is more vulnerable to over-fitting.

## 5. Benchmark Results

### 5.1. Overview of results

In this section we show the performance of the channel-out network on several image classification benchmarks. For all the channel-out networks used in our experiments, the channel selection function is the $\max(\cdot)$ function. We run tests on CIFAR-10, CIFAR-100 [11] and STL-10 [4], significantly outperforming the state-of-the-art results on the last two datasets.

Our implementation is built on top of the efficient convolution CUDA kernels developed by Alex Krizhevsky [11]. Most experiments are done using a Tesla C2050 GPU. Due to the time limit, we had to control the size of our networks, and we could only put a limited effort on optimizing the hyper-parameters (totally relying on manual trials). Despite of these constraints, we still managed to get state-of-the-art performance on CIFAR-100 (63.4%) and STL-10 (69.5%), indicating the great potential of channel-out networks on difficult classification tasks. We believe that we can further improve our current result of channel-out networks on CIFAR-10 (86.80%) if parameters are better tuned.

We did not put our effort on developing optimized channel-out networks for easier image classification tasks, such as MNIST [12] and SVHN [15] (digit images) since we believe these tasks will not demonstrate the strengths of channel-out networks, which is better at encoding a large amount of highly variant patterns.

### 5.2. CIFAR-10

The CIFAR-10 dataset[11] consists of $32 \times 32 \times 3$ small color images of 10 object classes. There are 5000 training images and 1000 test images for each class. There are significant variations in the images regarding shape and pose, and the images are not well centered. Previous experiments on CIFAR-10 typically take two methodologies: with or without data augmentation. To better compare the generalization power of different models, we focus on the non-augmented version, namely using the whole images without any translation, flipping or rotation. However, we did perform ZCA whitening using the *make_cifar10_gcn_whitened.py* code in the pylearn2 repository [6].

The network used for CIFAR-10 experiment consists of

| Method | Precision |
|---|---|
| Maxout+Dropout [7] | 88.32% |
| Channel-out+Dropout | **86.80%** |
| CNN+Spearmint [17] | 85.02% |
| Stochastic Pooling [21] | 84.87% |

Table 1: Best methods on CIFAR-10

3 convolutional channel-out layers, followed by a fully connected channel-out layer, and then the softmax layer. The best model has 64-192-192 filters for the corresponding convolutional layers, and 1210 nodes in the fully connected layer. Each convolutional layer consists of a linear convolution portion, a max pooling portion and the channel-out portion. The fully-connected layer has a fully-connected linear portion and a channel-out portion. The channel-out group sizes are set as 2-2-2-5 for corresponding layers. Dropout with probability 0.5 is applied to the inputs of layers 2,3,4, and dropout with probability 0.2 is applied to the input of the first layer - the image itself. Here we would like to thank Goodfellow, etc. for providing an opensource for the maxout code in pylearn2, since many of our hyper-parameters for the CIFAR-10 dataset were tuned with the help of the .yaml configuration files in the pylearn2 repository.

The best test precision of the channel-out network is 86.80%, which is not as good as the state-of-the-art reported in [7] using the maxout network (88.32%), but is better than any of the other previous methods (as far as we know). The best results on CIFAR-10 without data augmentation are summarized in Table 1. We believe that the channel-out performance could be further improved if we use a larger network and the hyper-parameters are better tuned.

Aware of the fact that the performance can be highly dependent on hyper-parameter settings, we implemented our own code for maxout networks and compared them with our model using two different control settings: similar number of parameters or similar number of feature maps. For the former case, the maxout network is of size 96-256-256-1210 (number of filters/feature maps in each layer). A maxout network of this size has a similar total number of parameters (weights) as our channel-out network. For the latter case, it is of size 64-192-192-1210, i.e. having exactly same number of feature maps as the channel-out network. Note that this results in significantly less parameters in the maxout network. The results are shown in Table 2.

We can see that the channel-out network has nearly same performance as a maxout network with similar number of parameters, but is significantly better than a maxout network with same number of feature maps. We argue that the second comparison is reasonable in the sense that although having different numbers of parameters, their implementations require the same number of multiplication/addition operations, resulting in a similar training time. The faster version of the channel-out network is still under construction, which will take advantage of the structured sparseness of outputs of channel-out groups. We predict that it will be at least significantly faster than the maxout network with a similar number of parameters.

| Network | Precision |
|---|---|
| Channel-out+Dropout | **86.80%** |
| Maxout with similar number of parameters | 86.73% |
| Maxout with same number of feature maps | 84.35% |

Table 2: Comparison between channel-out and maxout with different settings

| Method | Precision |
|---|---|
| Channel-out+Dropout | **63.41%** |
| Maxout+Dropout [7] | 61.43% |
| Stochastic Pooling [21] | 57.49% |
| Receptive Field Learning [9] | 54.23% |

Table 3: Best methods on CIFAR-100

### 5.3. CIFAR-100

The CIFAR-100 dataset [11] is similar to CIFAR-10, but with 100 classes. There are 500 training images and 100 test images for each class. Both the larger number of labels and the smaller training sets make the task much more difficult than the CIFAR-10 dataset.

The channel-out network developed for the CIFAR-100 has 3 convolutional layers followed by a fully connected layer and the softmax layer, with feature maps 80-176-176-1210. The channel-out group sizes are 2-2-2-5. Similar to the CIFAR-10 experiment, we applied 0.2 dropout probability to the input layer and 0.5 dropout to all other layers. Images are pre-whitened and when presented to the network each time, they are horizontally flipped with probability 0.5. The test set precision was 63.41%, improving the current state-of-the-art by nearly 2 percentage points. Table 3 shows the best results on CIFAR-100.

Motivated by the better performance on the CIFAR-100, we perform another experiment to illustrate the fact that channel-out networks are better at harder tasks where the data patterns show more variances. We extract 10, 20, 50, 100 classes from the original dataset (both training and testing) to generate 4 training-testing pairs. We train a channel-out and a maxout network with a similar number of parameters for each of the four classification tasks. The performance results are shown in Figure 8. Note that to facilitate faster experimentation, we have deliberately used much smaller networks for this series of experiments.

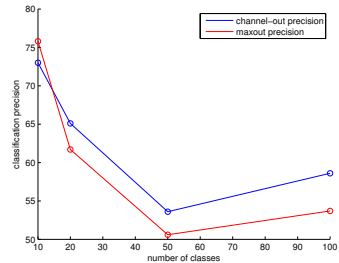

Figure 8: Comparison of channel-out and maxout on 4 tasks of different difficult levels: channel-out does better on harder tasks.

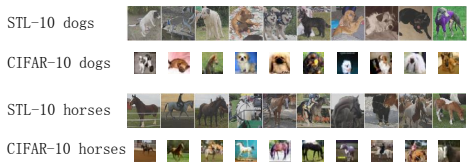

Figure 9: Example images in STL-10 and CIFAR-10 datasets: STL-10 dataset has more variations.

The results justify our claim that channel-out is not best suited for the case when there is relatively a small amount of information that needs to be stored (the 10-class task). In such a case, interference between patterns is less of a problem. As the patterns become more complex, channel-out surpasses maxout due to its better implementation of sparse pathway encoding. Notice the unusual fact that both methods perform better for the 100-class task than for the 50-class task. This may be due to uneven distribution of difficulty among the data. The first 50 classes seem to be more difficult to discriminate.

### 5.4. STL-10

STL-10 [4] contains images from 10 classes, typically with more variant shapes and more complex backgrounds than CIFAR-10 images. Figure 9 shows the comparison between STL-10 and CIFAR-10 images. The original images are of size $96 \times 96 \times 3$. There are 500 training and 800 test images for each class. There are additional 100,000 unlabeled images for unsupervised learning.

Our best channel-out network for STL-10 consists of three convolutional layers followed a fully-connected layer and the softmax layer, with feature maps 64-176-256-1212. The channel-out group sizes are set as 4-4-4-4. Dropout is applied same way as before. To reduce the complexity, we down-sampled all images to $32 \times 32$. Images are pre-whitened, and horizontally flipped with probability 0.5 when presented to the network. We got the state-of-the-art test precision of 69.5%, which improves the current state-of-the-art by 5%. Best results on STL-10 are shown in Table 4. Note that in our experiments the network is trained with the training set only. The unsupervised corpus is only involved in the whitening process.

| Method | Precision |
|---|---|
| Channel-out+Dropout | **69.5%** |
| Hierarchical Matching Pursuit [2] | 64.5% |
| Discriminative Learning of SPN [5] | 62.3% |

Table 4: Best methods on STL-10

The good performance on STL-10 shows that channel-out network does a better job at discriminating more variant patterns, and ruling out the interference of complex backgrounds. This further demonstrates that information is stored more efficiently in a channel-out network.

## 6. Conclusions

We have introduced the concept of sparse pathway coding and argued that this can be a robust and efficient way for encoding categorical information in a deep network. Using sparse pathway encoding, the interference between conflicting patterns is mitigated, and therefore when combined with dropout, the network can utilize the network capacity in a more effective way. Along this direction we have proposed a novel class of deep networks, the channel-out networks. Our experiments show that channel-out networks perform very well on image classification tasks, especially for the harder tasks with more complex patterns. We believe that the concept of sparse pathway coding is well worth pursuing further for designing robust deep networks.

## 7. Acknowledgements

Upon finishing this paper, we found that a recent work from the IDSIA lab [19] proposed a similar model as the $\max(\cdot)$ version of the channel-out network. Our work was independently developed. Compared to their work, our model was motivated and analysed from a different perspective. We have also provided new theoretical and experimental results regarding the $\max(\cdot)$ channel-out model.

## References

[1] Y. Bengio. Learning deep architectures for AI. *Foundations and trends® in Machine Learning*, 2(1):1–127, 2009.
[2] L. Bo, X. Ren, and D. Fox. Unsupervised feature learning for rgb-d based object recognition. *ISER, June*, 2012.
[3] Y.-l. Boureau, Y. L. Cun, et al. Sparse feature learning for deep belief networks. In *Advances in neural information processing systems*, pages 1185–1192, 2007.
[4] A. Coates, A. Y. Ng, and H. Lee. An analysis of single-layer networks in unsupervised feature learning. In *International Conference on Artificial Intelligence and Statistics*, pages 215–223, 2011.
[5] R. Gens and P. Domingos. Discriminative learning of sum-product networks. In *Advances in Neural Information Processing Systems*, pages 3248–3256, 2012.
[6] I. J. Goodfellow, D. Warde-Farley, P. Lamblin, V. Dumoulin, M. Mirza, R. Pascanu, J. Bergstra, F. Bastien, and Y. Bengio. Pylearn2: a machine learning research library. *arXiv preprint arXiv:1308.4214*, 2013.
[7] I. J. Goodfellow, D. Warde-Farley, M. Mirza, A. Courville, and Y. Bengio. Maxout networks. *arXiv preprint arXiv:1302.4389*, 2013.
[8] G. E. Hinton, N. Srivastava, A. Krizhevsky, I. Sutskever, and R. R. Salakhutdinov. Improving neural networks by preventing co-adaptation of feature detectors. *arXiv preprint arXiv:1207.0580*, 2012.
[9] Y. Jia, C. Huang, and T. Darrell. Beyond spatial pyramids: Receptive field learning for pooled image features. In *Computer Vision and Pattern Recognition (CVPR), 2012 IEEE Conference on*, pages 3370–3377. IEEE, 2012.
[10] E. R. Kandel, J. H. Schwartz, T. M. Jessell, et al. *Principles of neural science*, volume 4. McGraw-Hill New York, 2000.
[11] A. Krizhevsky, I. Sutskever, and G. Hinton. ImageNet classification with deep convolutional neural networks. In *Advances in Neural Information Processing Systems 25*, pages 1106–1114, 2012.
[12] Y. LeCun, L. Bottou, Y. Bengio, and P. Haffner. Gradient-based learning applied to document recognition. *Proceedings of the IEEE*, 86(11):2278–2324, 1998.
[13] H. Lee, C. Ekanadham, and A. Ng. Sparse deep belief net model for visual area V2. In *Advances in neural information processing systems*, pages 873–880, 2007.
[14] E. Mendelson. *Introduction to mathematical logic*. CRC press, 1997.
[15] Y. Netzer, T. Wang, A. Coates, A. Bissacco, B. Wu, and A. Y. Ng. Reading digits in natural images with unsupervised feature learning. In *NIPS Workshop on Deep Learning and Unsupervised Feature Learning*, volume 2011, 2011.
[16] C. Poultney, S. Chopra, Y. L. Cun, et al. Efficient learning of sparse representations with an energy-based model. In *Advances in neural information processing systems*, pages 1137–1144, 2006.
[17] J. Snoek, H. Larochelle, and R. P. Adams. Practical Bayesian optimization of machine learning algorithms. *arXiv preprint arXiv:1206.2944*, 2012.
[18] N. Srivastava. *Improving neural networks with dropout*. PhD thesis, University of Toronto, 2013.
[19] R. K. Srivastava, J. Masci, S. Kazerounian, F. Gomez, and J. Schmidhuber. Compete to compute. *technical report*, 2013.
[20] L. Wan, M. Zeiler, S. Zhang, Y. L. Cun, and R. Fergus. Regularization of neural networks using dropconnect. In *Proceedings of the 30th International Conference on Machine Learning (ICML-13)*, pages 1058–1066, 2013.
[21] M. D. Zeiler and R. Fergus. Stochastic pooling for regularization of deep convolutional neural networks. *arXiv preprint arXiv:1301.3557*, 2013.

# Appendix

## Proof of the universal approximation theorem

***Theorem***: *Any piece-wise continuous function defined on a compact domain in Euclidean space can be approximated arbitrarily well by a* $\max(\cdot)$ *two-layer channel-out network with one hidden channel-out group with finite number of candidate nodes. I.e., denote the target function to be approximated as* $T(\mathbf{x})$, $\forall \epsilon > 0$, *we can find a two-layer channel-out network with one hidden channel-out group that implements a function* $\hat{T}(\cdot)$ *such that*

$$\int \left| T(\mathbf{x}) - \hat{T}(\mathbf{x}) \right|^2 \mathbf{dx} < \epsilon$$

*Here by a "piece-wise continuous function" we refer to a function that is constituted of finite number of continuous segments.*

**Proof**

1) It's obvious that any (continuous) convex piece-wise linear function defined on a compact domain in Euclidean space can be expressed as a max-of-linear function $\max(\mathbf{Wx})$.

2) Select an arbitrary infinite number series that is positive, (strictly) monotonically increasing and bounded $\{g_1, g_2, g_3, ...\}$. Suppose $0 < g_i < G$.

3) We inductively construct a prototype function $P(\cdot)$, which is piece-wise linear, continuous and convex, defined on a hypercube centered at the origin and covering the domain of the target function.

We first segment the domain of $P(\cdot)$ ($n$-dimensional) into lattices with corner coordinates as integer multiples of some $\delta > 0$, i.e., the corner coordinates of a lattice can be written as $\{[k_1\delta, k_2\delta, ..., k_n\delta], [(k_1+1)\delta, k_2\delta, ..., k_n\delta], ..., [(k_1+1)\delta, (k_2+1)\delta, ..., (k_n+1)\delta]\}$ (totally $2^n$ corner points). For convenience, we refer to such lattice as $Lattice(k_1, k_2, ..., k_n)$. In each $Lattice(k_1, k_2, ..., k_n)$, $P(\cdot)$ is defined as a linear function plus a constant shift $c > 0$:

$$P(\mathbf{x}) = f_{(k_1, k_2, ..., k_n)}(\cdot) + c \tag{2}$$

$$c > 0, \mathbf{x} \in Lattice(k_1, k_2, ..., k_n)$$

The linear segments $f_{(k_1, k_2, ..., k_n)}(\cdot)$ are defined inductively. For a lattice where $\forall i, k_i > 0$, it is defined as following:

$$f_{(1,1,...,1)}(0, 0, ..., 0) = 0 \tag{3}$$

$$f_{(k_1, k_2, ..., k_n)}((k_1-1)\delta, (k_2-1)\delta, ..., (k_n-1)\delta) = f_{(k_1-1, k_2-1, ..., k_n-1)}((k_1-1)\delta, (k_2-1)\delta, ..., (k_n-1)\delta) \tag{4}$$

And the gradient in this lattice is defined as

$$\triangledown f_{(k_1, k_2, ..., k_n)}(\mathbf{x}) = (g_{k_1}, g_{k_2}, ..., g_{k_n}) \tag{5}$$

For all other lattices, the regional function is defined symmetrically with respect to its all-positive counterpart:

$$f_{(k_1, k_2, ..., k_n)}(\mathbf{x}) = f_{(|k_1|, |k_2|, ..., |k_n|)}(\mathbf{x}) \tag{6}$$

It's easy to verify that $P(\cdot)$ constructed in this way is convex and piece-wise continuous. Thus it can be expressed as a max-linear function $\max(\mathbf{Wx})$. Furthermore, $P(\mathbf{x}) \geq c$. Assign the linear weights $\mathbf{W}$ to the input weights of the channel-out group. The channel selection function is then constructed as $\arg\max(\mathbf{Wx})$.

4) Properly scale each $f(\cdot) + c$ to match the target function $T(\cdot)$. Define the approximation

$$\hat{T}(\mathbf{x}) = \gamma_{(k_1, k_2, ..., k_n)} P(\mathbf{x}), \text{ if } \mathbf{x} \in Lattice(k_1, k_2, ..., k_n) \tag{7}$$

where
$$\gamma_{(k_1,k_2,...,k_n)} = \frac{T((k_1-1)\delta, (k_2-1)\delta, ..., (k_n-1)\delta)}{f_{(k_1,k_2,...,k_n)}((k_1-1)\delta, (k_2-1)\delta, ..., (k_n-1)\delta) + c} \qquad (8)$$

Since $T(\cdot)$ is bounded, $\gamma_{(k_1,k_2,...,k_n)}$ is also bounded, say $\forall (k_1, k_2, ..., k_n), |\gamma_{(k_1,k_2,...,k_n)}| < r$. If $T$ is continuous in $Lattice(k_1, k_2, ..., k_n)$, then $\forall \epsilon_0 > 0$, we can choose $\delta$ such that

$$\left| \max_{\mathbf{x} \in Lattice(k_1,k_2,...,k_n)} T(\mathbf{x}) - \min_{\mathbf{x} \in Lattice(k_1,k_2,...,k_n)} T(\mathbf{x}) \right| < \epsilon_0 \qquad (9)$$

Since $P(\cdot)$ is piece-wise linear, we also know that

$$\left| \max_{\mathbf{x} \in Lattice(k_1,k_2,...,k_n)} P(\mathbf{x}) - \min_{\mathbf{x} \in Lattice(k_1,k_2,...,k_n)} P(\mathbf{x}) \right| \leq \sqrt{(g_{k_1}^2 + g_{k_2}^2 + ... + g_{k_n}^2) \cdot n\delta^2} < nG\delta \qquad (10)$$

and therefore

$$\left| \max_{\mathbf{x} \in Lattice(k_1,k_2,...,k_n)} \hat{T}(\mathbf{x}) - \min_{\mathbf{x} \in Lattice(k_1,k_2,...,k_n)} \hat{T}(\mathbf{x}) \right| < \left|\gamma_{(k_1,k_2,...,k_n)}\right| nG\delta < rnG\delta \qquad (11)$$

Hence
$$\left| T(\mathbf{x}) - \hat{T}(\mathbf{x}) \right| < \epsilon_0 + rnG\delta \qquad (12)$$

Since $T$ has a limit number of continuous segments, the set of discontinuous points has zero measure. Let $M$ be the number of non-continuous lattices, then
$$\lim_{\delta \to 0} M\delta^n = 0 \qquad (13)$$

Since $T$ must be bounded, suppose $\max_{\mathbf{x}} T(\mathbf{x}) - \min_{\mathbf{x}} T(\mathbf{x}) < B$, then in these non-continuous lattices

$$\left| T(\mathbf{x}) - \hat{T}(\mathbf{x}) \right| < B + rnG\delta \qquad (14)$$

Suppose the total volume of $Domain(T)$ is $V$, In summary we'll have

$$\int \left| T(\mathbf{x}) - \hat{T}(\mathbf{x}) \right|^2 d\mathbf{x} < (V - M\delta^n)(\epsilon_0 + rnG\delta)^2 + M\delta^n(B + rnG\delta)^2 \qquad (15)$$

and with (13) we have
$$\lim_{\delta \to 0} \int \left| T(\mathbf{x}) - \hat{T}(\mathbf{x}) \right|^2 d\mathbf{x} = 0 \qquad (16)$$

i.e. the function constructed as $\hat{T}(\cdot)$ can approximate $T(\cdot)$ to arbitrary precision (in $l_2$ sense).

Let K be the total number of lattices, construct the output weights of the channel-out network as $\mathbf{\Gamma} = (\gamma_1, \gamma_2, ..., \gamma_K)$ (where each $\gamma_i$ is some $\gamma_{(k_1,k_2,...,k_n)}$ previously defined). We can easily see that the channel-out network implements the function

$$\hat{T}(\mathbf{x}) = \sum_{i=1}^{K} \gamma_i \cdot \mathbf{I}_{\{i = \arg\max(\mathbf{W}\mathbf{x})\}} \cdot (\mathbf{W}\mathbf{x})_i \qquad (17)$$

where $(\mathbf{W}\mathbf{x})_i$ denotes the $i^{th}$ component of vector $\mathbf{W}\mathbf{x}$.

Q.E.D.